\documentclass{article}

\PassOptionsToPackage{numbers, compress}{natbib}

\usepackage[final]{tackling_climate_workshop_style}

\usepackage[utf8]{inputenc} 
\usepackage[T1]{fontenc}    
\usepackage{hyperref}       
\usepackage{url}            
\usepackage{booktabs}       
\usepackage{threeparttable}
\usepackage{enumitem}      
\usepackage{amsmath}        
\usepackage{amsfonts}       
\usepackage{nicefrac}       
\usepackage{microtype}      
\usepackage{xcolor}         
\usepackage{tikz}
\usetikzlibrary{positioning,arrows.meta,fit,calc,backgrounds}
\usepackage{todonotes}      

\newcommand{\result}[2]{\expandafter\gdef\csname res:#1\endcsname{#2}}
\newcommand{\R}[1]{\ifcsname res:#1\endcsname\csname res:#1\endcsname\else---\fi}
\newcommand{\B}[1]{\ifcsname res:best:#1\endcsname\textbf{\R{#1}}\else\R{#1}\fi}
\newcommand{\N}[1]{\ifcsname res:n.#1\endcsname\csname res:n.#1\endcsname\,(\csname res:covpct.#1\endcsname)\else---\fi}
\IfFileExists{results.tex}{
\result{cov.min}{54}   
\result{cov.max}{70}   
\result{cov.gc.wp}{0.55}
\result{cov.gc.na}{0.79}
\result{cov.pangu.wp}{0.54}
\result{cov.pangu.na}{0.79}
\result{cov.hres.wp}{0.61}
\result{cov.hres.na}{0.92}

\result{persist.none.wp.track.6}{38.28}
\result{persist.none.wp.track.12}{85.85}
\result{persist.none.wp.track.18}{141.37}
\result{persist.none.wp.track.24}{202.32}
\result{persist.none.wp.vmax.6}{5.37}
\result{persist.none.wp.vmax.12}{10.55}
\result{persist.none.wp.vmax.18}{15.41}
\result{persist.none.wp.vmax.24}{19.33}
\result{persist.none.na.track.6}{42.07}
\result{persist.none.na.track.12}{105.26}
\result{persist.none.na.track.18}{183.60}
\result{persist.none.na.track.24}{260.55}
\result{persist.none.na.vmax.6}{3.24}
\result{persist.none.na.vmax.12}{6.70}
\result{persist.none.na.vmax.18}{9.65}
\result{persist.none.na.vmax.24}{12.35}

\result{te.gc.wp.track.6}{51.06}
\result{te.gc.wp.track.12}{46.54}
\result{te.gc.wp.track.18}{52.44}
\result{te.gc.wp.track.24}{57.03}
\result{te.gc.wp.vmax.6}{22.44}
\result{te.gc.wp.vmax.12}{25.30}
\result{te.gc.wp.vmax.18}{26.63}
\result{te.gc.wp.vmax.24}{28.90}
\result{te.gc.na.track.6}{46.30}
\result{te.gc.na.track.12}{54.90}
\result{te.gc.na.track.18}{60.58}
\result{te.gc.na.track.24}{69.05}
\result{te.gc.na.vmax.6}{22.93}
\result{te.gc.na.vmax.12}{24.92}
\result{te.gc.na.vmax.18}{26.88}
\result{te.gc.na.vmax.24}{28.77}

\result{te.pangu.wp.track.6}{53.58}
\result{te.pangu.wp.track.12}{53.07}
\result{te.pangu.wp.track.18}{61.89}
\result{te.pangu.wp.track.24}{62.07}
\result{te.pangu.wp.vmax.6}{22.71}
\result{te.pangu.wp.vmax.12}{24.70}
\result{te.pangu.wp.vmax.18}{26.19}
\result{te.pangu.wp.vmax.24}{26.88}
\result{te.pangu.na.track.6}{46.57}
\result{te.pangu.na.track.12}{59.02}
\result{te.pangu.na.track.18}{63.02}
\result{te.pangu.na.track.24}{62.54}
\result{te.pangu.na.vmax.6}{23.54}
\result{te.pangu.na.vmax.12}{24.82}
\result{te.pangu.na.vmax.18}{26.67}
\result{te.pangu.na.vmax.24}{27.22}

\result{te.hres.wp.track.6}{59.41}
\result{te.hres.wp.track.12}{63.18}
\result{te.hres.wp.track.18}{72.50}
\result{te.hres.wp.track.24}{79.75}
\result{te.hres.wp.vmax.6}{15.08}
\result{te.hres.wp.vmax.12}{15.69}
\result{te.hres.wp.vmax.18}{16.17}
\result{te.hres.wp.vmax.24}{16.50}
\result{te.hres.na.track.6}{49.85}
\result{te.hres.na.track.12}{56.14}
\result{te.hres.na.track.18}{69.36}
\result{te.hres.na.track.24}{74.83}
\result{te.hres.na.vmax.6}{15.05}
\result{te.hres.na.vmax.12}{14.85}
\result{te.hres.na.vmax.18}{15.51}
\result{te.hres.na.vmax.24}{15.49}

\result{ours.gc.wp.track.6}{28.47}
\result{ours.gc.wp.track.12}{35.59}
\result{ours.gc.wp.track.18}{43.80}
\result{ours.gc.wp.vmax.6}{3.68}
\result{ours.gc.wp.vmax.12}{5.94}
\result{ours.gc.wp.vmax.18}{7.46}
\result{ours.gc.wp.vmax.24}{8.93}
\result{ours.gc.wp.track.24}{48.68}
\result{ours.gc.na.track.6}{30.41}
\result{ours.gc.na.track.12}{42.24}
\result{ours.gc.na.track.18}{51.52}
\result{ours.gc.na.track.24}{61.53}
\result{ours.gc.na.vmax.6}{3.31}
\result{ours.gc.na.vmax.12}{5.55}
\result{ours.gc.na.vmax.18}{7.00}
\result{ours.gc.na.vmax.24}{8.29}
\result{ours.pangu.wp.track.6}{29.48}
\result{ours.pangu.wp.track.12}{39.63}
\result{ours.pangu.wp.track.18}{47.76}
\result{ours.pangu.wp.track.24}{51.11}
\result{ours.pangu.wp.vmax.6}{3.75}
\result{ours.pangu.wp.vmax.12}{6.23}
\result{ours.pangu.wp.vmax.18}{8.26}
\result{ours.pangu.wp.vmax.24}{10.15}
\result{ours.hres.wp.track.6}{30.14}
\result{ours.hres.wp.track.12}{40.88}
\result{ours.hres.wp.track.18}{51.77}
\result{ours.hres.wp.track.24}{62.25}
\result{ours.hres.wp.vmax.6}{3.56}
\result{ours.hres.wp.vmax.12}{5.87}
\result{ours.hres.wp.vmax.18}{7.35}
\result{ours.hres.wp.vmax.24}{8.77}

\result{wnc.ours.bt.track.6}{33.55}
\result{wnc.ours.bt.track.12}{43.46}
\result{wnc.ours.bt.track.18}{51.12}
\result{wnc.ours.bt.track.24}{58.70}
\result{wnc.ours.bt.vmax.6}{4.20}
\result{wnc.ours.bt.vmax.12}{5.65}
\result{wnc.ours.bt.vmax.18}{6.69}
\result{wnc.ours.bt.vmax.24}{7.80}

\result{wnc.native.bt.track.6}{43.10}
\result{wnc.native.bt.track.12}{46.71}
\result{wnc.native.bt.track.18}{52.36}
\result{wnc.native.bt.track.24}{60.84}
\result{wnc.native.bt.vmax.6}{7.88}
\result{wnc.native.bt.vmax.12}{8.19}
\result{wnc.native.bt.vmax.18}{8.62}
\result{wnc.native.bt.vmax.24}{9.35}

\result{wnc.ours.tcv.track.6}{35.11}
\result{wnc.ours.tcv.track.12}{42.05}
\result{wnc.ours.tcv.track.18}{49.95}
\result{wnc.ours.tcv.track.24}{58.00}
\result{wnc.ours.tcv.vmax.6}{3.94}
\result{wnc.ours.tcv.vmax.12}{5.34}
\result{wnc.ours.tcv.vmax.18}{6.44}
\result{wnc.ours.tcv.vmax.24}{7.29}

\result{wnc.native.tcv.track.6}{37.33}
\result{wnc.native.tcv.track.12}{42.01}
\result{wnc.native.tcv.track.18}{49.40}
\result{wnc.native.tcv.track.24}{60.36}
\result{wnc.native.tcv.vmax.6}{7.95}
\result{wnc.native.tcv.vmax.12}{8.23}
\result{wnc.native.tcv.vmax.18}{8.71}
\result{wnc.native.tcv.vmax.24}{9.49}

\result{wnc.n.bt.track.6}{810}
\result{wnc.n.bt.track.12}{774}
\result{wnc.n.bt.track.18}{743}
\result{wnc.n.bt.track.24}{710}

\result{wnc.n.bt.vmax.6}{602}
\result{wnc.n.bt.vmax.12}{592}
\result{wnc.n.bt.vmax.18}{579}
\result{wnc.n.bt.vmax.24}{564}

\result{wnc.n.tcv.track.6}{639}
\result{wnc.n.tcv.track.12}{637}
\result{wnc.n.tcv.track.18}{633}
\result{wnc.n.tcv.track.24}{623}

\result{wnc.n.tcv.vmax.6}{586}
\result{wnc.n.tcv.vmax.12}{572}
\result{wnc.n.tcv.vmax.18}{559}
\result{wnc.n.tcv.vmax.24}{543}

\result{wnc.storms.bt.6}{29}
\result{wnc.storms.bt.24}{28}

\result{wnc.storms.tcv.6}{28}
\result{wnc.storms.tcv.24}{27}

\result{wnc.seederr.mean}{20.54}
\result{wnc.seederr.median}{11.08}

\result{best:wnc.ours.bt.track.6}{1}
\result{best:wnc.ours.bt.track.12}{1}
\result{best:wnc.ours.bt.track.18}{1}
\result{best:wnc.ours.bt.track.24}{1}

\result{best:wnc.ours.bt.vmax.6}{1}
\result{best:wnc.ours.bt.vmax.12}{1}
\result{best:wnc.ours.bt.vmax.18}{1}
\result{best:wnc.ours.bt.vmax.24}{1}

\result{best:wnc.ours.tcv.track.6}{1}
\result{best:wnc.native.tcv.track.12}{1}
\result{best:wnc.native.tcv.track.18}{1}
\result{best:wnc.ours.tcv.track.24}{1}

\result{best:wnc.ours.tcv.vmax.6}{1}
\result{best:wnc.ours.tcv.vmax.12}{1}
\result{best:wnc.ours.tcv.vmax.18}{1}
\result{best:wnc.ours.tcv.vmax.24}{1}

\result{n.gc.wp.6}{562}  \result{covpct.gc.wp.6}{62}
\result{n.gc.wp.12}{514}  \result{covpct.gc.wp.12}{58}
\result{n.gc.wp.18}{484}  \result{covpct.gc.wp.18}{57}
\result{n.gc.wp.24}{458}  \result{covpct.gc.wp.24}{55}
\result{n.pangu.wp.6}{562}  \result{covpct.pangu.wp.6}{62}
\result{n.pangu.wp.12}{510}  \result{covpct.pangu.wp.12}{58}
\result{n.pangu.wp.18}{481}  \result{covpct.pangu.wp.18}{56}
\result{n.pangu.wp.24}{450}  \result{covpct.pangu.wp.24}{54}
\result{n.hres.wp.6}{640}  \result{covpct.hres.wp.6}{70}
\result{n.hres.wp.12}{575}  \result{covpct.hres.wp.12}{65}
\result{n.hres.wp.18}{538}  \result{covpct.hres.wp.18}{63}
\result{n.hres.wp.24}{506}  \result{covpct.hres.wp.24}{61}
\result{n.int.gc.wp.6}{524}  \result{covpct.int.gc.wp.6}{70}
\result{n.int.gc.wp.12}{484}  \result{covpct.int.gc.wp.12}{67}
\result{n.int.gc.wp.18}{456}  \result{covpct.int.gc.wp.18}{66}
\result{n.int.gc.wp.24}{432}  \result{covpct.int.gc.wp.24}{65}
\result{n.int.pangu.wp.6}{527}  \result{covpct.int.pangu.wp.6}{71}
\result{n.int.pangu.wp.12}{487}  \result{covpct.int.pangu.wp.12}{67}
\result{n.int.pangu.wp.18}{459}  \result{covpct.int.pangu.wp.18}{67}
\result{n.int.pangu.wp.24}{431}  \result{covpct.int.pangu.wp.24}{65}
\result{n.int.hres.wp.6}{591}  \result{covpct.int.hres.wp.6}{79}
\result{n.int.hres.wp.12}{542}  \result{covpct.int.hres.wp.12}{75}
\result{n.int.hres.wp.18}{511}  \result{covpct.int.hres.wp.18}{74}
\result{n.int.hres.wp.24}{479}  \result{covpct.int.hres.wp.24}{72}
\result{n.gc.na.6}{187}  \result{covpct.gc.na.6}{86}
\result{n.gc.na.12}{176}  \result{covpct.gc.na.12}{84}
\result{n.gc.na.18}{158}  \result{covpct.gc.na.18}{80}
\result{n.gc.na.24}{151}  \result{covpct.gc.na.24}{79}
\result{n.pangu.na.6}{184}  \result{covpct.pangu.na.6}{85}
\result{n.pangu.na.12}{175}  \result{covpct.pangu.na.12}{83}
\result{n.pangu.na.18}{160}  \result{covpct.pangu.na.18}{81}
\result{n.pangu.na.24}{150}  \result{covpct.pangu.na.24}{79}
\result{n.hres.na.6}{207}  \result{covpct.hres.na.6}{95}
\result{n.hres.na.12}{198}  \result{covpct.hres.na.12}{94}
\result{n.hres.na.18}{185}  \result{covpct.hres.na.18}{93}
\result{n.hres.na.24}{175}  \result{covpct.hres.na.24}{92}

\result{best:ours.gc.wp.track.6}{1}
\result{best:ours.gc.wp.track.12}{1}
\result{best:ours.gc.wp.track.18}{1}
\result{best:ours.gc.wp.track.24}{1}
\result{best:ours.gc.wp.vmax.6}{1}
\result{best:ours.gc.wp.vmax.12}{1}
\result{best:ours.gc.wp.vmax.18}{1}
\result{best:ours.gc.wp.vmax.24}{1}
\result{best:ours.pangu.wp.track.6}{1}
\result{best:ours.pangu.wp.track.12}{1}
\result{best:ours.pangu.wp.track.18}{1}
\result{best:ours.pangu.wp.track.24}{1}
\result{best:ours.pangu.wp.vmax.6}{1}
\result{best:ours.pangu.wp.vmax.12}{1}
\result{best:ours.pangu.wp.vmax.18}{1}
\result{best:ours.pangu.wp.vmax.24}{1}
\result{best:ours.hres.wp.track.6}{1}
\result{best:ours.hres.wp.track.12}{1}
\result{best:ours.hres.wp.track.18}{1}
\result{best:ours.hres.wp.track.24}{1}
\result{best:ours.hres.wp.vmax.6}{1}
\result{best:ours.hres.wp.vmax.12}{1}
\result{best:ours.hres.wp.vmax.18}{1}
\result{best:ours.hres.wp.vmax.24}{1}

\result{abl.lmf2.ir.track.6}{34.86}
\result{abl.lmf2.ir.track.12}{49.42}
\result{abl.lmf2.ir.track.18}{60.26}
\result{abl.lmf2.ir.track.24}{69.22}
\result{abl.lmf2.ir.vmax.6}{3.39}
\result{abl.lmf2.ir.vmax.12}{5.17}
\result{abl.lmf2.ir.vmax.18}{6.37}
\result{abl.lmf2.ir.vmax.24}{7.45}
\result{abl.lmf2.noir.track.6}{35.60}
\result{abl.lmf2.noir.track.12}{50.82}
\result{abl.lmf2.noir.track.18}{61.74}
\result{abl.lmf2.noir.track.24}{71.37}
\result{abl.lmf2.noir.vmax.6}{3.38}
\result{abl.lmf2.noir.vmax.12}{5.11}
\result{abl.lmf2.noir.vmax.18}{6.43}
\result{abl.lmf2.noir.vmax.24}{7.60}
\result{abl.d.lmf2.track.6}{-0.74}
\result{abl.d.lmf2.track.12}{-1.40}
\result{abl.d.lmf2.track.18}{-1.48}
\result{abl.d.lmf2.track.24}{-2.15}
\result{abl.d.lmf2.vmax.6}{+0.01}
\result{abl.d.lmf2.vmax.12}{+0.06}
\result{abl.d.lmf2.vmax.18}{-0.06}
\result{abl.d.lmf2.vmax.24}{-0.15}
\result{abl.n.6}{1815}
\result{abl.n.12}{1758}
\result{abl.n.18}{1702}
\result{abl.n.24}{1646}
\result{abl.nint.6}{1510}
\result{abl.nint.12}{1468}
\result{abl.nint.18}{1424}
\result{abl.nint.24}{1378}
\result{best:abl.lmf2.ir.track.12}{1}
\result{best:abl.lmf2.ir.track.18}{1}
\result{best:abl.lmf2.ir.track.24}{1}
\result{best:abl.lmf2.ir.track.6}{1}
\result{best:abl.lmf2.ir.vmax.18}{1}
\result{best:abl.lmf2.ir.vmax.24}{1}
\result{best:abl.lmf2.noir.vmax.12}{1}
\result{best:abl.lmf2.noir.vmax.6}{1}

}{}
\IfFileExists{dataset_numbers.tex}{
\result{data.train.samples}{28{,}962}
\result{data.train.storms}{801}
\result{data.train.spill}{1{,}108}

\result{data.val2018.samples}{1{,}171}
\result{data.val2018.storms}{34}

\result{data.test2019.samples}{1{,}069}
\result{data.test2019.storms}{32}

\result{data.wp2020.samples}{746}
\result{data.wp2020.storms}{25}

\result{data.test1920.samples}{1{,}815}
\result{data.test1920.storms}{57}

\result{data.test2025.samples}{867}
\result{data.test2025.storms}{29}

}{}

\title{TC-Next: Zero-Shot Multimodal Cyclone Forecasting}

\author{%
  Zhe Wang\thanks{Equal contribution.}  \\
  Durham University\\
  Durham, United Kingdom \\
  \texttt{zhe@hostspace.ai} \\
  \And
  Sijie Chen\footnotemark[1] \\
  Carnegie Mellon University\\
  Pittsburgh, PA 15213 \\
  \texttt{sijieche@andrew.cmu.edu} \\
  \And
  Yiming Luo \\
  Carnegie Mellon University\\
  Pittsburgh, PA 15213 \\
  \texttt{yimingl1@andrew.cmu.edu} \\
  \AND
  Daehyun Kim \\
  Seoul National University\\
  Seoul, South Korea \\
  \texttt{daehyun@snu.ac.kr} \\
  \And
  Chien-Yi Chang \\
  Durham University\\
  Durham, United Kingdom \\
  \texttt{chien-yi.chang@durham.ac.uk} \\
}

\begin{document}

\maketitle

\vspace{-5mm}
\begin{abstract}

We present TropicalCycloneNext (TC-Next), a multimodal deep learning model that forecasts tropical cyclone track and intensity at $6$--$24$\,h leads by leveraging
a foundation model's forecast of atmospheric kinematic and thermodynamic fields, along with GridSat infrared satellite imagery.
Trained only on GraphCast forecasts over the Western Pacific (WP), yet reliant only on generic atmospheric variables, TC-Next on GraphCast lowers track error by $15$--$44\%$ and intensity error by a factor of $3$--$6$ relative to the rule-based tracker TempestExtremes; applied without retraining to Pangu-Weather and IFS HRES forecasts, it stays ahead on both. Applied zero-shot to the generic weather fields of WeatherNext Cyclones on the 2025 WP season, it attains lower average intensity error at every lead time, and lower or comparable average track error, than that model's specialized direct tracker in a deterministic comparison. Ablations show that the satellite branch lowers track error at every lead time and intensity error at longer leads. TC-Next thus gives a common, independently verifiable basis for comparing the AI forecasts that inform cyclone-warning and climate-adaptation decisions as best available science.

\vspace{-2mm}

\end{abstract}

\vspace{-3mm}
\section{Introduction}
\vspace{-3mm}

Tropical cyclones (TCs) are among the deadliest climate hazards, a threat accelerating under a warming climate \citep{emanuel2005, knutson2019}. Mass evacuations depend on accurate track and intensity forecasts, such as those that evacuated over two million people ahead of the $145$\,kt Super Typhoon Ragasa in 2025 \citep{ragasa2025}. 
Because adaptation is meant to be guided by the best available science \citep{paris,icj2025}, improving these forecasts and being able to tell which is best is a concrete form of climate adaptation.
Recent advances in deep learning have produced weather foundation models, including deterministic ones such as GraphCast \citep{graphcast}, Pangu-Weather \citep{pangu}, and Aurora \citep{aurora}, as well as probabilistic ones such as GenCast \citep{gencast} and FGN \citep{fgn}, that forecast the global atmospheric states, matching or outperforming operational numerical weather prediction (NWP) while running orders of magnitude faster. While they reproduce TC tracks well \citep{graphcast, gencast}, they struggle with intensity. When intensity is read off their fields using TC trackers like TempestExtremes \citep{tempestextremes}, the results are no better than persistence at short leads \citep{tcbench}. 
Furthermore, the ERA5 reanalysis \citep{era5} they are trained on carries a $12$--$15$\,m/s intensity error against best-track data \citep{knapp2010ibtracs} due to resolution constraints and assimilation processes \citep{fuxitc}.

\vspace{-0.3em}
Prior work addresses this issue in three main ways. First, WeatherNext Cyclones (WN-C) \citep{wnc} extends the underlying weather model with TC-specific feature channels trained using best-track labels and extracts track and intensity through a specialized heuristic direct tracker. This attains state-of-the-art intensity, but at the cost of leaving the tracker and the TC head locked to the specialized model. Second, Fuxi-TC downscales the atmospheric variables resolution to resolve the storm's tight inner core \citep{fuxitc} by using a diffusion model to map coarse 0.25° FuXi forecasts to 0.1° WRF~\citep{wrf}-simulated fields. However, its intensity skill is below its teacher model, WRF-0.1. Third, a number of end-to-end models that do not leverage foundation models have been proposed in recent years \citep[e.g.,][]{hurricast, cyclonemae,owzp}. Although these models are relatively successful at intensity prediction, they cannot match the track skill achieved by foundation models. Furthermore, no existing approach achieves zero-shot transfer across diverse forecast models while also leveraging satellite observations of the storm core.

\vspace{-0.3em}
We introduce \textbf{TropicalCycloneNext} (TC-Next), a multimodal deep learning model designed to predict TC track and intensity around Western Pacific (WP) basin. TC-Next uses a foundation model's forecast fields, which supply the large-scale steering flow and thermodynamic environmental conditions that are known affect TC intensity, together with high-resolution (0.07°) GridSat satellite imagery~\citep{gridsat} of the storm's recent history, which provides convective structures associated with the storm. Our contributions are three-fold:
\textbf{1. State-of-the-Art Zero-Shot Transfer:}\footnote{We define \emph{zero-shot transfer} as replacing the training-time forecast
source $F=\mathrm{GraphCast}$ (the publicly released $0.25^\circ$, $37$-level checkpoint trained on ERA5 1979--2017) with an unseen source $F'$ at inference
without retraining or modifying any other input.
The only exception is the comparison against WN-C in Table~\ref{tab:wnc}: there, IFS HRES analysis frames replace the ERA5 history frames to match the analysis distribution WN-C is fine-tuned on, and the TC~Vitals block further replaces the IBTrACS anchors with real-time TC~Vitals estimates.} We conduct comprehensive zero-shot out-of-distribution (OOD) evaluations across both AI-based and numerical forecasting models. Despite being trained solely on GraphCast, TC-Next achieved state-of-the-art transfer performance: it outperformed the specialized direct tracker of the current SOTA (WN-C) in a deterministic setup (Table~\ref{tab:wnc}), and consistently surpassed the TempestExtremes baseline in both track and intensity errors across multiple AI foundation models as well as the HRES numerical model at all lead times, lowering track error by $15$--$49\%$ and intensity error by a factor of $2$--$6$ (Table~\ref{tab:wp}).
\textbf{2. Multimodal Efficacy:} Ablation studies on the satellite infrared (IR) branch demonstrate that multimodal inputs lower track error at every lead time and improve intensity error from $+18$\,h onward. The benefit increases with lead time, peaking at $+24$\,h (Table~\ref{tab:abl} in Appendix~\ref{app:setup}).
\textbf{3. Cross-Model Portability:} 
TC-Next requires from the forecast source only generic atmospheric variables
(Table~\ref{tab:channels}; SST is persisted from the analysis).
Hence, the same trained model can be applied across forecast sources, providing a common learned tracker for evaluating TC forecast skill.

\vspace{-3mm}
\section{Methodology}
\vspace{-3mm}

\paragraph{Problem formulation.}
\label{sec:problem}

We forecast TC track and intensity at lead times
$\mathcal{L}=\{6,12,18,24\}\,\mathrm{h}$. Given observations through
time $t$, a network $f_\theta$ emits, for each lead $\ell$, raw center
and extent updates $\mathbf{a}_{\ell},\mathbf{s}_{\ell}\in\mathbb{R}^{2}$
and an intensity increment $u_{\ell}\in\mathbb{R}$:
\begin{equation}
\label{eq:task}
\big\{(\mathbf{a}_{\ell},\mathbf{s}_{\ell},u_{\ell})\big\}_{\ell\in\mathcal{L}}
=
f_{\theta}\!\left(
E_{t-24:t}, S_{t-24:t}, \hat{E}^{F}_{t+6:t+24},
\mathbf{c}_{t-24:t}, \mathbf{v}_{t-24:t}
\right).
\end{equation}
The center and intensity updates are applied autoregressively from time $t$:
\begin{equation}
\label{eq:update}
\hat{\mathbf{c}}_{t+\ell}=\hat{\mathbf{c}}_{t+\ell-6}+\delta\tanh(\mathbf{a}_{\ell}),\qquad
\hat{v}_{t+\ell}=\hat{v}_{t+\ell-6}+u_{\ell},
\end{equation}
where $\delta=4^\circ$ bounds the motion per step along each axis;
$\hat{\mathbf{c}}_{t+\ell}$ and $\hat{v}_{t+\ell}$ are the predicted
center and intensity at $t+\ell$.
Here, $E_{t-24:t}\in\mathbb{R}^{5\times H\times W\times C}$ holds
five historical atmospheric frames with $H\times W=280\times420$ grid
cells (a $70^\circ\times105^\circ$ domain at $0.25^\circ$) and $C=12$
channels (Table~\ref{tab:channels});
$S_{t-24:t}\in\mathbb{R}^{5\times256\times256\times2}$ holds five IR
frames, each a $16^\circ$ window of brightness temperature and its
validity mask; $\mathbf{c}_{t-24:t}\in\mathbb{R}^{5\times2}$ denotes
historical centers; and $\mathbf{v}_{t-24:t}\in\mathbb{R}^{5}$
is the historical intensity. The four future atmospheric frames
$\hat{E}^{F}_{t+6:t+24}\in\mathbb{R}^{4\times H\times W\times C}$ are
forecasts from a weather foundation model $F$ initialized at $t$. 
The extent $\mathbf{s}_\ell$ updates an auxiliary box around the
storm (from \texttt{USA\_R34} wind radius) that lets the loss judge a
position miss by both distance and storm size; see Eq.~(\ref{eq:decode}) for its update rule.
The history window has is fixed at five steps, but past fixes are optional: only the anchor fix
$(\mathbf{c}_t, v_t)$ must be observed, so a forecast can issue from a
storm's first fix.\footnote{History steps preceding the storm's first
record are zero-filled and masked, in training and at
inference alike.}

\vspace{-3mm}
\paragraph{Data.} 
\label{sec:data}



To train TC-Next, we built a dataset combining atmospheric fields with IR brightness temperatures near $11\,\mu\mathrm{m}$ from GridSat-B1 and storm state from IBTrACS. The atmospheric fields come from ERA5 reanalysis for the storm's observed history and from GraphCast forecasts for the lead times ahead.
From each atmospheric source we crop a fixed, WP-basin-centered $70^\circ\times105^\circ$ domain; $0.07^\circ$ IR is resampled to $0.0625^\circ$ and cropped to a $16^\circ$ window centered on the storm's observed position at each history step. Storm centers follow IBTrACS's best track, and intensity is the $1$-minute maximum sustained wind (\texttt{USA\_WIND}).

A training sample is anchored to a single IBTrACS fix at time $t$. Over the five $6$-hourly steps ending at $t$, it gathers ERA5, IR, and TC positions and intensities, masking any steps that precede the storm's first record. It also includes the four GraphCast forecasts initialized at $t$. As targets, we use the storm's positions and intensities from $t{+}6$ to $t{+}24$\,h, masking any leads beyond its final record. Every in-domain fix that is followed by another in-domain fix $6$\,h later is a sample; storms that never reach tropical-storm strength (peak \texttt{USA\_SSHS} $<0$) are dropped. Details are in Appendix~\ref{app:data}. We partition data chronologically to prevent temporal leakage: training (1990--2017), validation (2018), and testing (2019--2020), comprising \R{data.train.samples}, \R{data.val2018.samples}, and \R{data.test1920.samples} samples from \R{data.train.storms}, \R{data.val2018.storms}, and \R{data.test1920.storms} storms, respectively. To avoid temporal overlap with WN-C's training data, we evaluate against it on a separate 2025 test set of \R{data.test2025.samples} samples from \R{data.test2025.storms} storms.
At evaluation, we score track and intensity separately at each lead time: track error is the mean great-circle distance (km), and intensity error is the mean absolute error (kt). Beyond GraphCast, we draw Pangu-Weather \citep{pangu} and IFS HRES \citep{ifs} forecasts from the WeatherBench2 archive \citep{weatherbench2}, and real-time storm estimates from TC~Vitals \citep{tcvitals}.

\vspace{-3mm}
\paragraph{Model.}
\label{sec:model}

TC-Next is a multimodal encoder-decoder architecture with $3.6$\,M parameters, trainable on a TPU v4-8 VM in less than 70 minutes. Its two encoders process atmospheric fields and satellite imagery at their respective input resolutions. A \emph{macro} CNN maps each atmospheric frame to a $2^\circ$ feature grid. A finer-stride \emph{micro} CNN processes a $16^\circ$ GridSat window centered on the observed storm position at each historical time step, together with its per-pixel validity mask, producing features on a $0.25^\circ$ grid.
The macro feature map goes into a region-proposal network \citep{faster_rcnn} (RPN) to predict a storm-centered proposal box and the extent, supervised directly against the center and wind-radii box at every history step, to give the macro CNN a gradient path that bypasses the recurrent encoder--decoder. RoIAlign \citep{mask_rcnn} then extracts geographically aligned $10^\circ$ windows around the proposed center from both feature maps, and low-rank multimodal fusion \citep{lmf} combines the resulting representations; see Eq.~(\ref{eq:fusion}).

\vspace{-0.3em}
An encoder LSTM processes the five fused historical representations and initializes a decoder LSTM. At each forecast lead $\ell$, the decoder extracts a $10^\circ$ window from the macro-encoded forecast frame at $t+\ell$, centered on its previous predicted storm position. Separate prediction heads produce the center and intensity updates of
Eq.~(\ref{eq:update}) together with an extent update $\mathbf{s}_{\ell}$
that scales the predicted box extent
$\hat{\mathbf{r}}_{t+\ell}\in\mathbb{R}^{2}$ (width and height):
\begin{equation}
\label{eq:decode}
\hat{\mathbf{r}}_{t+\ell}=\operatorname{clip}_{[2^\circ,10^\circ]}\!\Big(\hat{\mathbf{r}}_{t+\ell-6}\odot\exp\!\big(\tanh(\mathbf{s}_{\ell})/2\big)\Big).
\end{equation}
The decoder initializes $\hat{\mathbf{c}}_{t}$ and $\hat{\mathbf{r}}_{t}$
from the last RPN proposal at $t$, and sets $\hat{v}_{t}=v_t$. Each forecast
therefore updates an existing estimate of storm position, extent, and
intensity. The decoder then rolls forward autoregressively over the available forecast frames. Architectural and
training details, including the loss in Eq.~(\ref{eq:loss}), appear in
Appendix~\ref{app:model}.

\vspace{-3mm}
\section{Experiments and Results}
\vspace{-3mm}
We evaluate TC-Next in three settings: (1) against the direct tracker of WN-C on the WP 2025 season (Table~\ref{tab:wnc}); (2) against TempestExtremes \citep{tempestextremes}, configured as in TCBench \citep{tcbench}, on the pooled WP 2019--2020 seasons across three forecast sources (Table~\ref{tab:wp}); and (3) with its satellite infrared branch ablated (Table~\ref{tab:abl} in Appendix~\ref{app:setup}). TC-Next is trained solely on GraphCast fields; Pangu-Weather, IFS HRES, and WN-C are evaluated without retraining. Baseline configuration, sample matching, pooling, and scoring conventions are given in Appendix~\ref{app:setup}.

\begin{table}
\small
  \centering
  \caption{WP, 2025; TC-Next against the direct tracker of WN-C. Errors as defined in Section~\ref{sec:data}; bold marks the lower error in each column within an anchor block. Scored samples: Table~\ref{tab:wnc-n}.}
  \label{tab:wnc}
  \vspace{-0.4em}
  \setlength{\tabcolsep}{2pt}
  \begin{threeparttable}
  \begin{tabular}{ll rrrr rrrr}
    \toprule
    & & \multicolumn{4}{c}{Track $\downarrow$ (km)} & \multicolumn{4}{c}{Intensity $\downarrow$ ($V_{\max}$, kt)} \\
    \cmidrule(lr){3-6}\cmidrule(lr){7-10}
    Anchor record & TC model & $+6$\,h & $+12$\,h & $+18$\,h & $+24$\,h & $+6$\,h & $+12$\,h & $+18$\,h & $+24$\,h \\
    \midrule
    IBTrACS & \textbf{TC-Next (WN-C)}\textsuperscript{a} & \B{wnc.ours.bt.track.6} & \B{wnc.ours.bt.track.12} & \B{wnc.ours.bt.track.18} & \B{wnc.ours.bt.track.24} & \B{wnc.ours.bt.vmax.6} & \B{wnc.ours.bt.vmax.12} & \B{wnc.ours.bt.vmax.18} & \B{wnc.ours.bt.vmax.24} \\
    IBTrACS & WN-C direct tracker\textsuperscript{b} & \B{wnc.native.bt.track.6} & \B{wnc.native.bt.track.12} & \B{wnc.native.bt.track.18} & \B{wnc.native.bt.track.24} & \B{wnc.native.bt.vmax.6} & \B{wnc.native.bt.vmax.12} & \B{wnc.native.bt.vmax.18} & \B{wnc.native.bt.vmax.24} \\
    \addlinespace
    TC~Vitals & \textbf{TC-Next (WN-C)}\textsuperscript{a} & \B{wnc.ours.tcv.track.6} & \B{wnc.ours.tcv.track.12} & \B{wnc.ours.tcv.track.18} & \B{wnc.ours.tcv.track.24} & \B{wnc.ours.tcv.vmax.6} & \B{wnc.ours.tcv.vmax.12} & \B{wnc.ours.tcv.vmax.18} & \B{wnc.ours.tcv.vmax.24} \\
    TC~Vitals & WN-C direct tracker\textsuperscript{b} & \B{wnc.native.tcv.track.6} & \B{wnc.native.tcv.track.12} & \B{wnc.native.tcv.track.18} & \B{wnc.native.tcv.track.24} & \B{wnc.native.tcv.vmax.6} & \B{wnc.native.tcv.vmax.12} & \B{wnc.native.tcv.vmax.18} & \B{wnc.native.tcv.vmax.24} \\
    \bottomrule
  \end{tabular}
  \begin{tablenotes}[flushleft]
  \footnotesize
  \item[] \textsuperscript{a}Zero-shot: applied to WN-C without retraining, reading only WN-C's generic weather channels. \textsuperscript{b}The direct tracker derives its position and intensity from WN-C's dedicated TC feature channels.
  \end{tablenotes}
  \end{threeparttable}
  \vspace{-0.6em}
\end{table}

\vspace{-3mm}
\paragraph{TC-Next transfers zero-shot to WN-C and outperforms its direct tracker.} Reading only generic weather channels, our frozen model outperforms the WN-C direct tracker's single-member predictions on track and intensity. WN-C's ensemble members are scored individually against best track and averaged, so Table~\ref{tab:wnc} reflects expected single-member error, not ensemble-mean error. The IBTrACS block uses the same HRES atmospheric source and IBTrACS anchor convention used for WN-C fine-tuning: our track error is lower at every lead, and early-lead intensity error is roughly halved. 
The TC~Vitals block \citep{tcvitals} swaps the IBTrACS anchors for the operational, real-time ones.
Here, track performance remains on par: TC-Next leads at $+6$\,h and $+24$\,h and trails by under $0.6$\,km in between. Meanwhile, our intensity advantage widens, holding between $\R{wnc.ours.tcv.vmax.6}$ and $\R{wnc.ours.tcv.vmax.24}$\,kt against the direct tracker's $\R{wnc.native.tcv.vmax.6}$ to $\R{wnc.native.tcv.vmax.24}$\,kt across all leads.
These results suggest that the gains survive the shift to operational anchors; fine-tuning directly on real-time inputs is a natural next step.


\vspace{-3mm}
\paragraph{TC-Next outperforms TempestExtremes across all forecast sources and lead times.} On GraphCast, the source the model is trained on, track error is $15$--$44\%$ lower than TempestExtremes' (Table~\ref{tab:wp}), and intensity error is $3.2$--$6.1$ times lower. The margin survives zero-shot transfer: applied without retraining, TC-Next leads TempestExtremes on track by $18$--$45\%$ on Pangu-Weather and by $22$--$49\%$ on IFS HRES, and its intensity error stays below $11$\,kt on every source at every lead, whereas TempestExtremes' never falls below $15$\,kt. These results highlight that our learned model provides an alternative to rule-based detection across forecast sources.

\begin{table}
\small
  \centering
  \caption{WP, 2019--2020; TC-Next vs.\ TempestExtremes. Scored samples: Table~\ref{tab:wp-n}.}
  \label{tab:wp}
  \vspace{-0.4em}
  \setlength{\tabcolsep}{1.5pt}
  \begin{threeparttable}
  \begin{tabular}{ll rrrr rrrr}
    \toprule
    & & \multicolumn{4}{c}{Track $\downarrow$ (km)} & \multicolumn{4}{c}{Intensity $\downarrow$ ($V_{\max}$, kt)} \\
    \cmidrule(lr){3-6}\cmidrule(lr){7-10}
    Forecast source & TC model & $+6$\,h & $+12$\,h & $+18$\,h & $+24$\,h & $+6$\,h & $+12$\,h & $+18$\,h & $+24$\,h \\
    \midrule
    GraphCast & TempestExtremes & \B{te.gc.wp.track.6} & \B{te.gc.wp.track.12} & \B{te.gc.wp.track.18} & \B{te.gc.wp.track.24} & \B{te.gc.wp.vmax.6} & \B{te.gc.wp.vmax.12} & \B{te.gc.wp.vmax.18} & \B{te.gc.wp.vmax.24} \\
    GraphCast & \textbf{TC-Next}\textsuperscript{a} & \B{ours.gc.wp.track.6} & \B{ours.gc.wp.track.12} & \B{ours.gc.wp.track.18} & \B{ours.gc.wp.track.24} & \B{ours.gc.wp.vmax.6} & \B{ours.gc.wp.vmax.12} & \B{ours.gc.wp.vmax.18} & \B{ours.gc.wp.vmax.24} \\
    \addlinespace
    Pangu$^{\dagger}$ & TempestExtremes & \B{te.pangu.wp.track.6} & \B{te.pangu.wp.track.12} & \B{te.pangu.wp.track.18} & \B{te.pangu.wp.track.24} & \B{te.pangu.wp.vmax.6} & \B{te.pangu.wp.vmax.12} & \B{te.pangu.wp.vmax.18} & \B{te.pangu.wp.vmax.24} \\
    Pangu$^{\dagger}$ & \textbf{TC-Next (Pangu)}\textsuperscript{b} & \B{ours.pangu.wp.track.6} & \B{ours.pangu.wp.track.12} & \B{ours.pangu.wp.track.18} & \B{ours.pangu.wp.track.24} & \B{ours.pangu.wp.vmax.6} & \B{ours.pangu.wp.vmax.12} & \B{ours.pangu.wp.vmax.18} & \B{ours.pangu.wp.vmax.24} \\
    \addlinespace
    IFS HRES & TempestExtremes & \B{te.hres.wp.track.6} & \B{te.hres.wp.track.12} & \B{te.hres.wp.track.18} & \B{te.hres.wp.track.24} & \B{te.hres.wp.vmax.6} & \B{te.hres.wp.vmax.12} & \B{te.hres.wp.vmax.18} & \B{te.hres.wp.vmax.24} \\
    IFS HRES & \textbf{TC-Next (HRES)}\textsuperscript{b} & \B{ours.hres.wp.track.6} & \B{ours.hres.wp.track.12} & \B{ours.hres.wp.track.18} & \B{ours.hres.wp.track.24} & \B{ours.hres.wp.vmax.6} & \B{ours.hres.wp.vmax.12} & \B{ours.hres.wp.vmax.18} & \B{ours.hres.wp.vmax.24} \\
    \bottomrule
  \end{tabular}
  \begin{tablenotes}[flushleft]
  \footnotesize
  \item[] \textsuperscript{a}Source: the model was trained on this forecast source. \textsuperscript{b}Zero-shot: applied without retraining.
  \item[] \textsuperscript{$\dagger$}Pangu-Weather used 2019 for validation, so that year is not strictly held out.
  \end{tablenotes}
  \end{threeparttable}
  \vspace{-0.6em}
\end{table}

\vspace{-3mm}


\paragraph{The satellite infrared branch lowers track error at every lead and intensity error at longer leads, with a benefit that grows with lead time.} Table~\ref{tab:abl} (Appendix~\ref{app:setup}) ablates the satellite infrared branch on the full test set: the \emph{IR-ablated} variant is a separately trained model with the IR input zeroed in training and evaluation, all else fixed. The full model lowers track error at every lead, by $0.7$\,km at $+6$\,h rising to $2.2$\,km at $+24$\,h, and intensity error from $+18$\,h onward, by $0.06$ and $0.15$\,kt.

\vspace{-3mm}
\section{Conclusion}
\vspace{-3mm}


In this study, we introduce TC-Next, a multimodal DNN that predicts TC track and intensity using a foundation weather model's forecast fields and high-resolution satellite imagery of the storm's history. Because the model's forecast input relies exclusively on generic atmospheric variables, it can be applied across forecast sources without retraining. Evaluated on our WP dataset, TC-Next demonstrates robust zero-shot transfer capabilities across AI models, numerical systems, and real-time input configurations, consistently outperforming traditional rule-based detection methods. 
Ongoing work extends the dataset and zero-shot benchmark to additional basins, toward a global benchmark for evaluating forecast sources, as the best available science standard for adaptation requires.

\begin{ack}

We gratefully acknowledge Google for providing computing resources through the TPU Research Cloud (TRC) program. This work contains modified Copernicus Climate Change Service information (ERA5); neither the European Commission nor ECMWF is responsible for any use of the Copernicus information it contains. This work originated from a Carnegie Mellon University course project and was substantially developed at Durham University. We thank Ziping He for her initial contributions and Ralf Brown and Yonatan Bisk for their guidance. DK was supported by the Creative-Pioneering Researchers Program through Seoul National University and NRF grant RS-2025-02310080. ZW and CC were supported by the JusTN0W Strategic Research Fund through Durham University and ELEVATE-ProClima through the British Academy (EPG\textbackslash100778).
\end{ack}

%
%
%

\clearpage
\bibliographystyle{unsrtnat}
\bibliography{references}

@article{knapp2010ibtracs,
  title   = {The International Best Track Archive for Climate Stewardship ({IBTrACS}): Unifying Tropical Cyclone Data},
  author  = {Knapp, Kenneth R. and Kruk, Michael C. and Levinson, David H. and Diamond, Howard J. and Neumann, Charles J.},
  journal = {Bulletin of the American Meteorological Society},
  volume  = {91},
  number  = {3},
  pages   = {363--376},
  year    = {2010},
  doi     = {10.1175/2009BAMS2755.1}
}

@article{gridsat,
  title   = {Globally Gridded Satellite Observations for Climate Studies},
  author  = {Knapp, Kenneth R. and Ansari, Steve and Bain, Caroline L. and Bourassa, Mark A. and Dickinson, Michael J. and Funk, Chris and Helms, Chip N. and Hennon, Christopher C. and Holmes, Christopher D. and Huffman, George J. and Kossin, James P. and Lee, Hai-Tien and Loew, Alexander and Magnusdottir, Gudrun},
  journal = {Bulletin of the American Meteorological Society},
  volume  = {92},
  number  = {7},
  pages   = {893--907},
  year    = {2011},
  doi     = {10.1175/2011BAMS3039.1}
}

@article{era5,
  title   = {The {ERA5} global reanalysis},
  author  = {Hersbach, Hans and Bell, Bill and Berrisford, Paul and Hirahara, Shoji and Hor{\'a}nyi, Andr{\'a}s and Mu{\~n}oz-Sabater, Joaqu{\'i}n and Nicolas, Julien and Peubey, Carole and Radu, Raluca and Schepers, Dinand and others},
  journal = {Quarterly Journal of the Royal Meteorological Society},
  volume  = {146},
  number  = {730},
  pages   = {1999--2049},
  year    = {2020},
  doi     = {10.1002/qj.3803}
}

@article{emanuel2005,
  title   = {Increasing destructiveness of tropical cyclones over the past 30 years},
  author  = {Emanuel, Kerry},
  journal = {Nature},
  volume  = {436},
  number  = {7051},
  pages   = {686--688},
  year    = {2005},
  doi     = {10.1038/nature03906}
}

@article{knutson2019,
  title   = {Tropical Cyclones and Climate Change Assessment: Part {I}: Detection and Attribution},
  author  = {Knutson, Thomas and Camargo, Suzana J. and Chan, Johnny C. L. and Emanuel, Kerry and Ho, Chang-Hoi and Kossin, James and Mohapatra, Mrutyunjay and Satoh, Masaki and Sugi, Masato and Walsh, Kevin and Wu, Liguang},
  journal = {Bulletin of the American Meteorological Society},
  volume  = {100},
  number  = {10},
  pages   = {1987--2007},
  year    = {2019},
  doi     = {10.1175/BAMS-D-18-0189.1}
}

@techreport{ragasa2025,
  title       = {Post-Event Report: 2025 Western North Pacific Typhoon Ragasa},
  author      = {{Guy Carpenter}},
  institution = {Guy Carpenter \& Company, LLC},
  month       = oct,
  year        = {2025},
  note        = {\url{https://www.guycarp.com/insights/2025/10/post-event-report-2025-western-north-pacific-typhoon-rRagasa.html}}
}

@article{graphcast,
  title   = {Learning skillful medium-range global weather forecasting},
  author  = {Lam, Remi and Sanchez-Gonzalez, Alvaro and Willson, Matthew and Wirnsberger, Peter and Fortunato, Meire and Alet, Ferran and Ravuri, Suman and Ewalds, Timo and Eaton-Rosen, Zach and Hu, Weihua and Merose, Alexander and Hoyer, Stephan and Holland, George and Vinyals, Oriol and Stott, Jacklynn and Pritzel, Alexander and Mohamed, Shakir and Battaglia, Peter},
  journal = {Science},
  volume  = {382},
  number  = {6677},
  pages   = {1416--1421},
  year    = {2023}
}

@article{pangu,
  title   = {Accurate medium-range global weather forecasting with {3D} neural networks},
  author  = {Bi, Kaifeng and Xie, Lingxi and Zhang, Hengheng and Chen, Xin and Gu, Xiaotao and Tian, Qi},
  journal = {Nature},
  volume  = {619},
  pages   = {533--538},
  year    = {2023}
}

@article{gencast,
  title   = {Probabilistic weather forecasting with machine learning},
  author  = {Price, Ilan and Sanchez-Gonzalez, Alvaro and Alet, Ferran and Andersson, Tom R. and El-Kadi, Andrew and Masters, Dominic and Ewalds, Timo and Stott, Jacklynn and Mohamed, Shakir and Battaglia, Peter and Lam, Remi and Willson, Matthew},
  journal = {Nature},
  year    = {2025},
  note    = {GenCast; arXiv:2312.15796}
}

@article{fgn,
  title   = {Skillful joint probabilistic weather forecasting from marginals},
  author  = {Alet, Ferran and Price, Ilan and El-Kadi, Andrew and Masters, Dominic and Markou, Stratis and Andersson, Tom R. and Stott, Jacklynn and Lam, Remi and Willson, Matthew and Sanchez-Gonzalez, Alvaro and Battaglia, Peter},
  year    = {2025},
  note    = {FGN; arXiv:2506.10772}
}

@article{aurora,
  title   = {A foundation model for the {E}arth system},
  author  = {Bodnar, Cristian and Bruinsma, Wessel P. and Lucic, Ana and Stanley, Megan and Vaughan, Anna and Brandstetter, Johannes and Garvan, Patrick and Riechert, Maik and Weyn, Jonathan A. and Dong, Haiyu and Gupta, Jayesh K. and Thambiratnam, Kit and Archibald, Alexander T. and Wu, Chun-Chieh and Heider, Elizabeth and Welling, Max and Turner, Richard E. and Perdikaris, Paris},
  journal = {Nature},
  volume  = {641},
  year    = {2025},
  note    = {Aurora; arXiv:2405.13063}
}

@article{wnc,
  title   = {Operational Tropical Cyclone Forecasting with {AI}},
  author  = {Alet, Ferran and Andersson, Tom R. and Price, Ilan and Markou, Stratis and El-Kadi, Andrew and Masters, Dominic and Li, Amy and Merchant, Samier and Williams, Natalie and Thornton, Gregory and MacKay, Ken and Graham, Olivia and Uddin, Akib and Gaiarin, Ben and Shah, Devaja and Kruse, Elinor and Hogsett, Wallace and Zelinsky, David and Cangialosi, John and Martinez, Jonathan and Franklin, James and DeMaria, Mark and Musgrave, Kate and Bain, Caroline L. and Titley, Helen and Stott, Jacklynn and Lam, Remi and Bell, Aaron and Komarek, Paul and Willson, Matthew and Sanchez-Gonzalez, Alvaro and Battaglia, Peter},
  journal = {Nature},
  year    = {2026},
  doi     = {10.1038/s41586-026-10953-2},
  note    = {WeatherNext Cyclones}
}

@article{hurricast,
  title   = {Hurricane Forecasting: A Novel Multimodal Machine Learning Framework},
  author  = {Boussioux, L{\'e}onard and Zeng, Cynthia and Gu{\'e}nais, Th{\'e}o and Bertsimas, Dimitris},
  journal = {Weather and Forecasting},
  volume  = {37},
  number  = {6},
  pages   = {817--831},
  year    = {2022}
}

@article{owzp,
  title   = {Enhancing tropical cyclone track and intensity predictions with the {OWZP-Transformer} model},
  author  = {Lin, Zihao and Chu, Jung-Eun and Ham, Yoo-Geun},
  journal = {npj Artificial Intelligence},
  year    = {2025},
  doi     = {10.1038/s44387-025-00037-3}
}

@article{fuxitc,
  title   = {{FuXi-TC}: A generative framework integrating deep learning and physics-based models for improved tropical cyclone forecasts},
  author  = {Guo, Shan and Chen, Lei and Zhao, Yangyang and Lin, Yuetan and Niu, Zeyi and Zhang, Xinyan and Sun, Ziyao and Zhong, Xiaohui and Li, Hao},
  year    = {2025},
  note    = {arXiv:2508.16168}
}

@article{cyclonemae,
  title   = {{CycloneMAE}: A Scalable Multi-Task Learning Model for Global Tropical Cyclone Probabilistic Forecasting},
  author  = {Hang, Renlong and Xu, Zihao and Zhao, Jiuwei and Yu, Runling and Cheng, Leye and Liu, Qingshan},
  year    = {2026},
  note    = {arXiv:2604.12180}
}

@article{tempestextremes,
  title   = {{TempestExtremes} v2.1: a community framework for feature detection, tracking, and analysis in large datasets},
  author  = {Ullrich, Paul A. and Zarzycki, Colin M. and McClenny, Elizabeth E. and Pinheiro, Marielle C. and Stansfield, Alyssa M. and Reed, Kevin A.},
  journal = {Geoscientific Model Development},
  volume  = {14},
  number  = {8},
  pages   = {5023--5048},
  year    = {2021},
  doi     = {10.5194/gmd-14-5023-2021}
}

@article{tcbench,
  title   = {{TCBench}: A Benchmark for Tropical Cyclone Track and Intensity Forecasting at the Global Scale},
  author  = {Gomez, Milton and Ganesh S., Saranya and McGraw, Marie and Tam, Frederick Iat-Hin and Azizi, Ilia and Darmon, Samuel and Feldmann, Monika and Bourdin, Stella and Poulain-Auz{\'e}au, Louis and Camargo, Suzana J. and Lin, Jonathan and Chavas, Dan and Lee, Chia-Ying and Gupta, Ritwik and Jenney, Andrea and Beucler, Tom},
  year    = {2026},
  note    = {arXiv:2601.23268}
}

@article{tcvitals,
  title   = {An Analysis of {NCEP} Tropical Cyclone Vitals and Potential Effects on Forecasting Models},
  author  = {Trahan, Sam and Sparling, Lynn},
  journal = {Weather and Forecasting},
  volume  = {27},
  number  = {3},
  pages   = {744--756},
  year    = {2012},
  doi     = {10.1175/WAF-D-11-00063.1}
}

@article{weatherbench2,
  title   = {{WeatherBench} 2: A Benchmark for the Next Generation of Data-Driven Global Weather Models},
  author  = {Rasp, Stephan and Hoyer, Stephan and Merose, Alexander and Langmore, Ian and Battaglia, Peter and Russell, Tyler and Sanchez-Gonzalez, Alvaro and Yang, Vivian and Carver, Rob and Agrawal, Shreya and Chantry, Matthew and Ben Bouallegue, Zied and Dueben, Peter and Bromberg, Carla and Sisk, Jared and Barrington, Luke and Bell, Aaron and Sha, Fei},
  journal = {Journal of Advances in Modeling Earth Systems},
  volume  = {16},
  number  = {6},
  pages   = {e2023MS004019},
  year    = {2024},
  doi     = {10.1029/2023MS004019}
}

@inproceedings{faster_rcnn,
  title     = {Faster {R-CNN}: Towards Real-Time Object Detection with Region Proposal Networks},
  author    = {Ren, Shaoqing and He, Kaiming and Girshick, Ross and Sun, Jian},
  booktitle = {Advances in Neural Information Processing Systems},
  year      = {2015}
}

@inproceedings{mask_rcnn,
  title     = {Mask {R-CNN}},
  author    = {He, Kaiming and Gkioxari, Georgia and Doll{\'a}r, Piotr and Girshick, Ross},
  booktitle = {IEEE International Conference on Computer Vision (ICCV)},
  year      = {2017}
}

@inproceedings{lmf,
  title     = {Efficient Low-rank Multimodal Fusion with Modality-Specific Factors},
  author    = {Liu, Zhun and Shen, Ying and Lakshminarasimhan, Varun Bharadhwaj and Liang, Paul Pu and Zadeh, Amir and Morency, Louis-Philippe},
  booktitle = {Proceedings of the 56th Annual Meeting of the Association for Computational Linguistics (ACL)},
  year      = {2018}
}

@techreport{ifs,
  title       = {{IFS} Documentation {CY49R1} -- Part {IV}: Physical Processes},
  author      = {{ECMWF}},
  institution = {European Centre for Medium-Range Weather Forecasts},
  address     = {Reading, UK},
  year        = {2024},
  note        = {\url{https://www.ecmwf.int/en/elibrary/81626-ifs-documentation-cy49r1-part-iv-physical-processes}}
}

@techreport{wrf,
  title       = {A Description of the Advanced Research {WRF} Model Version 4.3},
  author      = {Skamarock, William C. and Klemp, Joseph B. and Dudhia, Jimy and Gill, David O. and Liu, Zhiquan and Berner, Judith and Wang, Wei and Powers, Jordan G. and Duda, Michael G. and Barker, Dale M. and Huang, Xiang-Yu},
  institution = {National Center for Atmospheric Research},
  address     = {Boulder, CO},
  number      = {NCAR/TN-556+STR},
  year        = {2021},
  doi         = {10.5065/1dfh-6p97}
}

@misc{icj2025,
  author       = {{International Court of Justice}},
  title        = {Obligations of States in respect of Climate Change},
  howpublished = {Advisory Opinion of 23 July 2025, General List No.~187},
  year         = {2025},
  url          = {https://www.icj-cij.org/sites/default/files/case-related/187/187-20250723-adv-01-00-en.pdf}
}

@misc{paris,
  author       = {{United Nations}},
  title        = {Paris Agreement},
  howpublished = {Adopted 12 December 2015, entered into force 4 November 2016, 3156 U.N.T.S. 79},
  year         = {2015},
  url          = {https://unfccc.int/sites/default/files/english_paris_agreement.pdf}
}

\clearpage
\appendix

\section{Data Details}
\label{app:data}

Table~\ref{tab:data} gives the sample and storm counts per split and the rule behind them; Table~\ref{tab:channels} lists the channels extracted per frame.

\paragraph{Data sources and preprocessing.} 
GraphCast's forecasts use the same $0.25^\circ$ grid and the same variables as the ERA5 history frames; both are read into one array with no regridding.
GridSat-B1 is the NOAA climate data record of intercalibrated geostationary imagery, native to a $0.07^\circ$ grid at $3$-h cadence; we take its single infrared-window channel at the 00, 06, 12 and 18\,UTC frames that coincide with the reanalysis cadence and resample it by bilinear interpolation. Pixels without a valid source observation, and whole frames absent from the record ($43$ in 1990--2018, $25$ in 2020), are zero-filled and carried as a per-pixel validity mask that the model receives as a second input channel. Each of the $12$ atmospheric channels is normalized by its mean and standard deviation, computed in one pass over all five frames of every one of the \R{data.train.samples} training samples, the statistics are then fixed for validation, testing, and every forecast source. Infrared brightness temperature is scaled by fixed constants, $(T_b-250\,\mathrm{K})/40\,\mathrm{K}$, and multiplied by the validity mask. 

The best-track records are post-season reanalyses for 1990--2020, whereas the 2025 records remain provisional. Storms that never reach tropical-storm strength are excluded, and fixes outside the domain are discarded.
Storm positions are kept in one coordinate system throughout: each best-track position (longitude wrapped to $0$--$360^\circ$, so the domain stays continuous across the antimeridian) is mapped linearly onto the $0.25^\circ$ latitude--longitude grid and expressed as a fractional (row, column) location in $[0,1]^2$ of the domain zero-padded to $288\times448$ cells ($72^\circ\times112^\circ$ extent). The model reads, proposes, and predicts positions in these units; multiplying by that extent converts them to the degrees used by the box loss, and forecasts are mapped back to latitude and longitude only for scoring.
A fix with no reported wind is masked out of the intensity term only. The extent target is a square box centered on the best-track position whose side is twice the mean of the four quadrant $34$\,kt wind radii reported by JTWC (IBTrACS \texttt{USA\_R34}), converted from nautical miles to degrees and clipped so that the side lies in $[2^\circ,10^\circ]$. Where no radii are reported, the side is fixed at $4^\circ$. When a storm's record spans less than $24$\,h at the anchor time, the unavailable history steps are masked rather than dropping the sample.



\paragraph{Array dimensions and storage.} Atmospheric fields are cached as one file per initialization time $t$: five frames $\times\,280\times420\times12$ channels in half precision (about $14.1$\,MB), holding the ERA5 analysis at $t$ and the four GraphCast forecasts initialized at $t$ (leads $6$--$24$\,h). A sample anchored at $t$ reads the analysis frame from the five files at $t{-}24,\dots,t$ and the four forecast frames from the file at $t$. 
Each analysis frame is therefore stored once and reused by every sample whose history window covers it, about $1.5$ samples per file on average. Because history windows straddle split boundaries, we report the dataset as total files rather than per-split counts.
GridSat infrared frames are stored separately, one per $6$-hourly step ($1120\times1680$ at resampled $0.0625^\circ$, compressed, about $1.7$\,MB each).

\paragraph{Dataset statistics.} The WP atmospheric cache (ERA5 and GraphCast) comprises $19{,}360$ files for 1990--2018, $770$ for 2019 and $491$ for 2020, or $20{,}621$ files and $271$\,GiB in total; the 2025 WN-C comparison adds a further $601$ files ($7.9$\,GiB). The infrared imagery totals $34.7$\,GiB for 1990--2020 and $1.00$\,GiB for 2025.

\begin{table}[!ht]
  \centering
  \caption{Dataset summary. The view is a fixed $70^\circ\times105^\circ$ domain at $0.25^\circ$ resolution spanning $5^\circ$S--$65^\circ$N and $95^\circ$E--$160^\circ$W. A sample is a $6$-hourly fix inside the view followed by another fix $6$\,h later, also inside it; storm counts are of the storms those samples come from. Because the view is a fixed geographic box rather than a basin mask, it also holds storms that IBTrACS codes to adjacent basins: \R{data.train.spill} of the \R{data.train.samples} training samples, most of them East Pacific. The test set pools two held-out seasons, 2019 (\R{data.test2019.samples} samples, \R{data.test2019.storms} storms) and 2020 (\R{data.wp2020.samples} samples, \R{data.wp2020.storms} storms). The 2025 counts come from the per-basin file \texttt{IBTrACS.WP.v04r01.nc} (snapshot of 2026-08-18) and are snapshot-dependent while the 2025 records remain provisional.}
  \label{tab:data}
  \footnotesize
  \setlength{\tabcolsep}{6pt}
  \begin{tabular}{lrrrrrrrr}
    \toprule
    & \multicolumn{2}{c}{Train (1990--2017)} & \multicolumn{2}{c}{Val (2018)} & \multicolumn{2}{c}{Test (2019--2020)} & \multicolumn{2}{c}{Test (2025)} \\
    \cmidrule(lr){2-3}\cmidrule(lr){4-5}\cmidrule(lr){6-7}\cmidrule(lr){8-9}
    Basin & Samples & Storms & Samples & Storms & Samples & Storms & Samples & Storms \\
    \midrule
    WP & \R{data.train.samples} & \R{data.train.storms} & \R{data.val2018.samples} & \R{data.val2018.storms} & \R{data.test1920.samples} & \R{data.test1920.storms} & \R{data.test2025.samples} & \R{data.test2025.storms} \\
    \bottomrule
  \end{tabular}
\end{table}

\begin{table}[!ht]
  \caption{Atmospheric channels extracted per frame from the analysis (history) and GraphCast (forecast). Only three pressure levels are used, chosen for the environmental controls on TC intensity: $850$\,hPa (low-level circulation and moisture), $500$\,hPa (mid-tropospheric environmental flow and core ascent), and $200$\,hPa (outflow; paired with $850$\,hPa it yields the deep-layer vertical wind shear). Sea-surface temperature is not forecast by GraphCast and is taken from the ERA5 analysis at initialization time $t$ in every forecast frame.}
  \label{tab:channels}
  \centering
  \small
  \begin{tabular}{llc}
    \toprule
    Level & Variables & Channels \\
    \midrule
    Surface & Mean sea-level pressure; $10$\,m wind ($u$, $v$) & 3 \\
    $850$\,hPa & Wind ($u$, $v$); specific humidity & 3 \\
    $500$\,hPa & Geopotential; vertical velocity ($\omega$) & 2 \\
    $200$\,hPa & Temperature; wind ($u$, $v$) & 3 \\
    Surface (analysis at $t$) & Sea-surface temperature & 1 \\
    \bottomrule
  \end{tabular}
\end{table}

\section{Model and Training Details}
\label{app:model}

\paragraph{Architecture.} A \emph{macro} CNN encodes each atmospheric frame with three stride-$2$ blocks of $32$, $64$, and $96$ channels, reducing the $0.25^\circ$ input to a $2^\circ$ feature map. Its input is the $12$ atmospheric channels of Table~\ref{tab:channels} plus four static planes: latitude, the sine and cosine of longitude, and the ERA5 land--sea mask. A finer-stride \emph{micro} CNN processes the $16^\circ$ GridSat window centered on the observed storm position at each history step, together with its per-pixel validity mask, mapping the resampled $0.0625^\circ$ IR input to a $0.25^\circ$ feature grid; the two maps then sit on scales RoIAlign can align, but the micro features are drawn from imagery four times finer than the environment fields.

A region-proposal network \citep{faster_rcnn} reads the macro map alongside a Gaussian prior rendered at the storm's known fix for that step and emits a box: a center offset from the prior, clamped to $2.5^\circ$, plus a side length in $[2^\circ,10^\circ]$. The proposal decides where the model looks: RoIAlign \citep{mask_rcnn} samples a fixed $10^\circ$ window from both feature maps at the proposed center ($8\times8$ points on the macro map, $16\times16$ on the micro map), and separate convolutional heads flatten each crop to $2048$ dimensions.

Write $z^{m}_{\tau}\in\mathbb{R}^{2048}$ for the crop vector of stream $m\in\{\mathrm{env},\mathrm{ir}\}$ at history step $\tau$. Each is projected to $128$ dimensions, layer-normalized with its own learned scale and offset, and extended by a constant:
\begin{equation}
\tilde{z}^{m}_{\tau}=\big[\mathrm{LN}_{m}(P_{m}z^{m}_{\tau});\,1\big]\in\mathbb{R}^{129}.
\end{equation}
The two streams are combined by low-rank multimodal fusion \citep{lmf},
\begin{equation}
\label{eq:fusion}
V_{\tau}=\sum_{r=1}^{R} w_{r}\,\big(W^{(r)}_{\mathrm{env}}\tilde{z}^{\mathrm{env}}_{\tau}\big)\odot\big(W^{(r)}_{\mathrm{ir}}\tilde{z}^{\mathrm{ir}}_{\tau}\big)+b,
\end{equation}
with $R=8$ and $\odot$ the element-wise product. The appended $1$ retains the unimodal terms, so Eq.~(\ref{eq:fusion}) is strictly more expressive than passing the concatenation of the two streams through a linear layer. The per-stream normalization is load-bearing: in a product each stream's gradient is scaled by the other, so without it the infrared branch is starved by the much larger environment signal.

The storm's own scalars are concatenated to $V_{\tau}$: the prior position, current intensity, availability flags, and a step index. The result passes through a two-layer MLP of width $256$ with ReLU, and an encoder LSTM of width $256$ reads the five resulting vectors in order; its final state initializes a decoder LSTM of the same width.
At each lead $\ell$, the decoder input is the $2048$-dimensional macro crop head applied to the RoIAlign window of the forecast frame at $t+\ell$, taken at the previous predicted center, concatenated with that center, the previous extent (divided by $10^\circ$), the previous intensity, and a one-hot lead index; a linear layer with ReLU maps this to $256$ dimensions before the LSTM. Two linear heads read the LSTM output: a four-dimensional box head giving $\mathbf{a}_{\ell}$ and $\mathbf{s}_{\ell}$ of Eqs.~(\ref{eq:update}) and~(\ref{eq:decode}), and a scalar head giving $u_{\ell}$. Both heads are initialized near zero, so the untrained model forecasts persistence. The IR data enters the forecast only through the encoder state: the decoder reads the forecast environment fields alone, since no IR imagery exists for future times.

\paragraph{Training.} 

The total loss has three terms of two kinds: a box loss, which scores position and extent, applied to both the forecast boxes and the history proposals; and an intensity loss, which scores maximum sustained wind.
Writing $b_{\ell}=(\mathbf{c}_{\ell},\mathbf{r}_{\ell})$ for the storm's box in degrees, the extent target $\mathbf{r}_{\ell}$ being the best-track gale-force ($34$\,kt) wind radius (R34, the mean of the four reported quadrants; a fixed $4^\circ$ where unreported), training minimizes
\begin{equation}
\label{eq:loss}
\begin{array}{c}
\mathcal{L}_{\mathrm{box}}\big(\mathcal{T}_{F}\big)+\beta\,\mathcal{L}_{\mathrm{int}}+\gamma\,\mathcal{L}_{\mathrm{box}}\big(\mathcal{T}_{H}\big),\\[3pt]
\mathcal{L}_{\mathrm{box}}(\mathcal{T})=\dfrac{1}{|\mathcal{T}|}\displaystyle\sum_{\ell\in\mathcal{T}}\Big[\lambda\,\|\hat{b}_{\ell}-b_{\ell}\|_{1}+1-\mathrm{GIoU}(\hat{b}_{\ell},b_{\ell})\Big],\qquad
\mathcal{L}_{\mathrm{int}}=\dfrac{1}{|\mathcal{T}_{v}|}\displaystyle\sum_{\ell\in\mathcal{T}_{v}}(\hat{v}_{\ell}-v_{\ell})^{2},
\end{array}
\end{equation}
with $\lambda=0.2$ and $\gamma=0.3$. Here $\mathcal{T}_{F}\subseteq\{t{+}6,\dots,t{+}24\}$ and $\mathcal{T}_{H}\subseteq\{t{-}24,\dots,t\}$ are the forecast and history steps at which the storm has a best-track fix, and $\mathcal{T}_{v}\subseteq\mathcal{T}_{F}$ those at which wind is also reported, so a storm with a partial record contributes only its observed fixes. For $\ell\in\mathcal{T}_{H}$, $\hat{b}_{\ell}$ is the region-proposal box at that step. The intensity weight $\beta$ opens at $20\%$ of its final value of $2.5$ and reaches it after one epoch, which lets the network localize before intensity competes for the shared trunk. We optimize with AdamW at a peak learning rate of $3\times10^{-4}$, warmed up over $300$ steps and cosine-decayed thereafter, with weight decay $10^{-4}$, gradients clipped to global norm $1$, and a batch of $32$ sharded data-parallel over a TPU v4-8. The model trains on the 1990--2017 split for $15$ epochs in about $65$ minutes; we keep the epoch with the best validation score on 2018, track error in km over $100$ plus intensity MAE in knots over $10$, which selects epoch $6$. No held-out season enters selection at any point. Under zero-shot transfer the input normalization statistics remain those of the training source; none are re-estimated.

\section{Experimental Setup Details}
\label{app:setup}

\paragraph{TempestExtremes.} For every forecast source we read the storm's position and intensity directly off its forecast fields with TempestExtremes \citep{tempestextremes}, using the detection parameters of TCBench \citep{tcbench}: a sea-level-pressure minimum is accepted only if the pressure rises by $200$\,Pa within $6.5^\circ$ great-circle distance (a closed low) and the $300$--$500$\,hPa thickness falls by $58.8$\,m$^{2}$\,s$^{-2}$ within $5.5^\circ$ (a warm core), with candidate nodes closer than $6.0^\circ$ merged. Intensity is measured at each node as the minimum sea-level pressure at the node itself and the maximum $10$\,m wind speed within $2.0^\circ$ great-circle distance. The baseline receives the same anchor as TC-Next. Where no node falls inside the gate the tracker returns no forecast: it produces a forecast for $\R{cov.min}$--$\R{cov.max}\%$ of samples depending on source, whereas TC-Next covers all samples by construction. All comparisons are therefore restricted to the samples both methods produce. Two caveats apply. First, the $2.0^\circ$ wind radius is the default of the standard TempestExtremes recipe; all baseline intensities are conditional on this choice. Second, a gridded wind maximum over a finite radius is not the same quantity as the $1$-minute sustained wind of the best-track record, so part of the baseline's intensity error is definitional rather than forecast error.

\paragraph{Forecast sources.} TC-Next is trained on GraphCast forecast fields only, together with ERA5 history frames, GridSat IR frames, and best-track anchors from IBTrACS. Every other input appears at evaluation time alone: the model is applied without retraining to the forecast fields of Pangu-Weather, IFS HRES, and WN-C, with the Pangu-Weather and IFS HRES fields taken from the WeatherBench2 archives \citep{weatherbench2}. Pangu-Weather emits no vertical velocity, so the $\omega_{500}$ input channel is zero-filled for its rows; TempestExtremes does not use this field and is unaffected. Table~\ref{tab:wnc} likewise substitutes inputs at evaluation time only. Throughout the table, IFS HRES analysis frames replace the ERA5 history, matching the analyses WN-C is fine-tuned on. TC~Vitals \citep{tcvitals} are the real-time storm estimates issued by the operational forecast centers. In the TC~Vitals block, they additionally replace the best-track anchors; that configuration therefore measures zero-shot robustness to the inputs available in real time, not the behavior of a retrained variant.

\paragraph{Scoring.} Table~\ref{tab:wp} pools the 2019 and 2020 seasons, each cell weighted by its own per-season matched count, and uses the $00$:$00$ and $12$:$00$\,UTC initialization times, the only times at which Pangu-Weather and IFS HRES are archived. Within each forecast source, all rows are scored on the samples that source's tracker matched, so each comparison is homogeneous. Table~\ref{tab:wp-n} counts, under the Track columns, the forecast samples for which TempestExtremes found a storm near the anchored position at that lead time (TC-Next forecasts every sample with a best-track target); under the Intensity columns, the subset of those matched samples whose best-track record includes the $1$-minute wind. In Tables~\ref{tab:wnc} and~\ref{tab:wp}, track and intensity are therefore scored on different samples: samples without a best-track $1$-minute wind (\texttt{USA\_WIND}) are dropped from intensity scoring only. In Table~\ref{tab:wnc}, WN-C is fine-tuned on IFS HRES analyses, so TC-Next is given HRES analysis frames in place of the ERA5 history to keep it on the same analysis distribution. The WN-C tracker can predict that a storm has dissipated while the IBTrACS still reports it; those rows are dropped for both models, so the IBTrACS block scores fewer than the $\R{data.test2025.samples}$ samples Table~\ref{tab:data} counts for 2025. Table~\ref{tab:wnc-n} lists the scored samples per lead. Each of WN-C's four ensemble members is scored separately and its errors averaged, rather than the ensemble being collapsed first: this measures the two models rather than the error cancellation an ensemble mean buys. TC-Next is trained against a deterministic backbone, and adapting it to an ensemble would require a different loss, which we leave to future work. Table~\ref{tab:abl} is scored on every held-out sample at all synoptic times; Table~\ref{tab:abl-n} lists those counts.

\begin{table}[!ht]
  \centering
  \caption{WP, 2019--2020; IR ablation study. \emph{Fed}: the full model. \emph{IR-ablated}: a separately trained model with the IR input zeroed in training and evaluation. A negative $\Delta$ means the IR branch lowers the error. Scored on every held-out sample, so the values are not directly comparable with Table~\ref{tab:wp}. Scored samples: Table~\ref{tab:abl-n}.}
  \label{tab:abl}
  \vspace{-0.4em}
  \setlength{\tabcolsep}{2.5pt}
  \begin{threeparttable}
  \begin{tabular}{l rrrr rrrr}
    \toprule
    & \multicolumn{4}{c}{Track $\downarrow$ (km)} & \multicolumn{4}{c}{Intensity $\downarrow$ ($V_{\max}$, kt)} \\
    \cmidrule(lr){2-5}\cmidrule(lr){6-9}
    IR branch & $+6$\,h & $+12$\,h & $+18$\,h & $+24$\,h & $+6$\,h & $+12$\,h & $+18$\,h & $+24$\,h \\
    \midrule
    Fed & \B{abl.lmf2.ir.track.6} & \B{abl.lmf2.ir.track.12} & \B{abl.lmf2.ir.track.18} & \B{abl.lmf2.ir.track.24} & \B{abl.lmf2.ir.vmax.6} & \B{abl.lmf2.ir.vmax.12} & \B{abl.lmf2.ir.vmax.18} & \B{abl.lmf2.ir.vmax.24} \\
    IR-ablated & \B{abl.lmf2.noir.track.6} & \B{abl.lmf2.noir.track.12} & \B{abl.lmf2.noir.track.18} & \B{abl.lmf2.noir.track.24} & \B{abl.lmf2.noir.vmax.6} & \B{abl.lmf2.noir.vmax.12} & \B{abl.lmf2.noir.vmax.18} & \B{abl.lmf2.noir.vmax.24} \\
    {\hspace{1em}\scriptsize $\Delta$ (fed $-$ IR-ablated)} & {\scriptsize$\R{abl.d.lmf2.track.6}$} & {\scriptsize$\R{abl.d.lmf2.track.12}$} & {\scriptsize$\R{abl.d.lmf2.track.18}$} & {\scriptsize$\R{abl.d.lmf2.track.24}$} & {\scriptsize$\R{abl.d.lmf2.vmax.6}$} & {\scriptsize$\R{abl.d.lmf2.vmax.12}$} & {\scriptsize$\R{abl.d.lmf2.vmax.18}$} & {\scriptsize$\R{abl.d.lmf2.vmax.24}$} \\
    \bottomrule
  \end{tabular}
  \end{threeparttable}
  \vspace{-0.6em}
\end{table}

\begin{table}[!ht]
  \centering
  \caption{Scored samples $n$ at each lead for Table~\ref{tab:wnc} (WP, 2025; TC-Next against the direct tracker of WN-C). Each row counts the samples on which both models of that anchor block are scored, namely those the WN-C tracker produced: where the WN-C tracker predicts that the storm has dissipated, the sample is dropped for both models. Intensity counts are lower because they also need a reported \texttt{USA\_WIND}.}
  \label{tab:wnc-n}
  \small
  \setlength{\tabcolsep}{4pt}
  \begin{tabular}{l rrrr rrrr}
    \toprule
    & \multicolumn{4}{c}{Track} & \multicolumn{4}{c}{Intensity} \\
    \cmidrule(lr){2-5}\cmidrule(lr){6-9}
    Anchor record & $+6$\,h & $+12$\,h & $+18$\,h & $+24$\,h & $+6$\,h & $+12$\,h & $+18$\,h & $+24$\,h \\
    \midrule
    IBTrACS & \R{wnc.n.bt.track.6} & \R{wnc.n.bt.track.12} & \R{wnc.n.bt.track.18} & \R{wnc.n.bt.track.24} & \R{wnc.n.bt.vmax.6} & \R{wnc.n.bt.vmax.12} & \R{wnc.n.bt.vmax.18} & \R{wnc.n.bt.vmax.24} \\
    TC~Vitals & \R{wnc.n.tcv.track.6} & \R{wnc.n.tcv.track.12} & \R{wnc.n.tcv.track.18} & \R{wnc.n.tcv.track.24} & \R{wnc.n.tcv.vmax.6} & \R{wnc.n.tcv.vmax.12} & \R{wnc.n.tcv.vmax.18} & \R{wnc.n.tcv.vmax.24} \\
    \bottomrule
  \end{tabular}
\end{table}

\begin{table}[!ht]
  \centering
  \caption{Scored samples $n$ at each lead for Table~\ref{tab:wp} (WP, pooled 2019--2020; TC-Next against TempestExtremes). Each row counts the samples on which both models of that forecast source are scored: the $00$:$00$ and $12$:$00$\,UTC samples (the only initialization times at which WeatherBench2 archives Pangu-Weather and IFS HRES forecasts) that have a best-track target at that lead and for which TempestExtremes found the storm; TC-Next forecasts every sample with a target, so the counts are limited only by TempestExtremes. Intensity counts are lower because they also need a reported \texttt{USA\_WIND}.}
  \label{tab:wp-n}
  \small
  \setlength{\tabcolsep}{4pt}
  \begin{tabular}{l rrrr rrrr}
    \toprule
    & \multicolumn{4}{c}{Track} & \multicolumn{4}{c}{Intensity} \\
    \cmidrule(lr){2-5}\cmidrule(lr){6-9}
    Forecast source & $+6$\,h & $+12$\,h & $+18$\,h & $+24$\,h & $+6$\,h & $+12$\,h & $+18$\,h & $+24$\,h \\
    \midrule
    GraphCast & \R{n.gc.wp.6} & \R{n.gc.wp.12} & \R{n.gc.wp.18} & \R{n.gc.wp.24} & \R{n.int.gc.wp.6} & \R{n.int.gc.wp.12} & \R{n.int.gc.wp.18} & \R{n.int.gc.wp.24} \\
    Pangu & \R{n.pangu.wp.6} & \R{n.pangu.wp.12} & \R{n.pangu.wp.18} & \R{n.pangu.wp.24} & \R{n.int.pangu.wp.6} & \R{n.int.pangu.wp.12} & \R{n.int.pangu.wp.18} & \R{n.int.pangu.wp.24} \\
    IFS HRES & \R{n.hres.wp.6} & \R{n.hres.wp.12} & \R{n.hres.wp.18} & \R{n.hres.wp.24} & \R{n.int.hres.wp.6} & \R{n.int.hres.wp.12} & \R{n.int.hres.wp.18} & \R{n.int.hres.wp.24} \\
    \bottomrule
  \end{tabular}
\end{table}

\begin{table}[!ht]
  \centering
  \caption{Scored samples $n$ at each lead for Table~\ref{tab:abl} (WP, 2019--2020; ablation of the IR branch). Both variants are scored on every held-out sample at all synoptic times, so the counts exceed those of Table~\ref{tab:wp-n} and the two tables are not directly comparable. Intensity counts are lower because they also need a reported \texttt{USA\_WIND}.}
  \label{tab:abl-n}
  \small
  \setlength{\tabcolsep}{4pt}
  \begin{tabular}{l rrrr rrrr}
    \toprule
    & \multicolumn{4}{c}{Track} & \multicolumn{4}{c}{Intensity} \\
    \cmidrule(lr){2-5}\cmidrule(lr){6-9}
    IR branch & $+6$\,h & $+12$\,h & $+18$\,h & $+24$\,h & $+6$\,h & $+12$\,h & $+18$\,h & $+24$\,h \\
    \midrule
    Fed and IR-ablated & \R{abl.n.6} & \R{abl.n.12} & \R{abl.n.18} & \R{abl.n.24} & \R{abl.nint.6} & \R{abl.nint.12} & \R{abl.nint.18} & \R{abl.nint.24} \\
    \bottomrule
  \end{tabular}
\end{table}

\end{document}